\documentclass[times,a4paper,twocolumn,final]{elsarticle}

\usepackage{amsfonts}       %
\usepackage{nicefrac}       %
\usepackage{microtype}      %
\usepackage{mathtools}
\usepackage{times}
\usepackage{latexsym}
\usepackage[utf8]{inputenc} %
\usepackage[T1]{fontenc}    %
\PassOptionsToPackage{hyphens}{url}\usepackage{hyperref}
\usepackage{url}            %
\usepackage{booktabs}       %
\usepackage{soul}
\usepackage{graphicx}
\usepackage{algorithm}
\usepackage{algorithmic}
\usepackage{microtype}
\usepackage{amsthm}
\usepackage{amssymb}
\usepackage{amsmath}
\usepackage{dsfont}
\usepackage{bbm}
\usepackage{xcolor}
\usepackage{multirow}

\usepackage{amsmath}
\usepackage[margin=1in]{geometry}

\usepackage{pgfplots}
\usepackage{filecontents}
\definecolor{tb_color_1}{RGB}{245,124,0}
\definecolor{tb_color_2}{RGB}{0,167,247}
\usepackage{tabularx}

\usepackage{style}
\usepackage{framed,multirow}
\usepackage{amssymb}
\usepackage{latexsym}
\usepackage{amsthm}
\usepackage{url}
\usepackage{xcolor}
\definecolor{newcolor}{rgb}{.8,.349,.1}
\usepackage{amsmath,amssymb}
\usepackage[utf8]{inputenc}
\usepackage{graphicx}%
\usepackage{braket}
\usepackage{tikz}
\usetikzlibrary{automata,arrows,arrows.meta,positioning,calc,shapes,trees}
\usepackage{bbm}%
\usepackage{subcaption}
\usepackage{relsize}
\usepackage{float}
\floatstyle{plaintop}
\restylefloat{table}
\usepackage{pgffor}
\usepackage[space]{grffile}
\usepackage{datatool}
\usepackage{csvsimple}
\usepackage{xspace}
\usepackage{etoolbox}
\usepackage{mathtools}
\usepackage{xargs}
\usepackage{hyperref}
\hypersetup{
    colorlinks,
    citecolor=black,
    filecolor=black,
    linkcolor=black,
    urlcolor=black
}
\usepackage{xhfill}
\usepackage[breakable, theorems, skins]{tcolorbox}
\usepackage[normalem]{ulem}
\usepackage{amsmath}
\usepackage{lipsum}

\DeclareMathOperator{\ME}{\mathds{E}}

\newcommand{\mbf}[1]{\mathbf{#1}}

\newcommand\numberthis[1][]{
    \refstepcounter{equation}
    \ifx#1\empty\else\label{eq:#1}\fi
    \tag{\theequation}
}

\newtheorem{theorem}{Theorem}[section]
\newtheorem{corollary}{Corollary}[theorem]
\newtheorem{lemma}[theorem]{Lemma}

\theoremstyle{definition}

\newcommand{\compmin}[1]{\min\{{#1}\}}
\newcommand{\compmax}[1]{\max\{{#1}\}}

\begin{document}

\twocolumn[
  \begin{@twocolumnfalse}

\begin{center}

{\LARGE \textbf{DGSAN: Discrete Generative Self-Adversarial Network}}

\vskip30pt

Ehsan Montahaei~~ Danial Alihosseini ~~ Mahdiyeh Soleymani Baghshah \\
\vskip10pt
  Sharif University of Technology,
  Tehran, 
  Iran\\
\texttt{ehsan.montahaei@gmail.com, dalihosseini@ce.sharif.edu, soleymani@sharif.edu} 

\end{center}

\section*{Abstract}
Although GAN-based methods have received many achievements in the last few years, they have not been entirely successful in generating discrete data. The most crucial challenge of these methods is the difficulty of passing the gradient from the discriminator to the generator when the generator outputs are discrete. Despite the fact that several attempts have been made to alleviate this problem, none of the existing GAN-based methods have improved the performance of text generation compared with the maximum likelihood approach in terms of both the quality and the diversity. In this paper, we proposed a new framework for generating discrete data by an adversarial approach in which there is no need to pass the gradient to the generator. The proposed method has an iterative manner in which each new generator is defined based on the last discriminator. It leverages the discreteness of data and the last discriminator to model the real data distribution implicitly.
Moreover, the method is supported with theoretical guarantees, and experimental results generally show the superiority of the proposed DGSAN method compared to the other popular or recent methods in generating discrete sequential data.

\vskip 1pt
\noindent \textbf{Keywords:} discrete data; generative model; self adversarial; sequential.

\vskip 20pt

\end{@twocolumnfalse}
]

\section{Introduction}

The early deep generative models that were utilized to generate sequential discrete data such as natural language were Recurrent Neural Networks (RNNs). 
However, RNN-based methods for discrete sequence generation that employ \textit{teacher forcing} approach for training suffer from the so-called \textit{exposure bias} problem \cite{ss-dis,sched-samp}. Moreover, the Generative Adversarial Networks (GANs) \cite{gan} were not entirely successful in generating sequential discrete data \cite{seqgan,gantutorial} despite their success in other domains, especially image generation. %

In the last years, various attempts were made to apply GANs to discrete domains, but some difficulties exist in the training process of these networks on discrete data. More precisely, in the discrete domains, passing the gradient to the generator is infeasible (due to the sampling at the output of the generator) during the training process \cite{seqgan,gantutorial}. To overcome this issue, \cite{seqgan} utilized Reinforcement Learning (RL) for the first time to this end.
Although RL provides a way to train the generator, it encounters some problems such as vast action space, the sparsity of the reward, and the high variance of the estimation. Since just receiving a scalar as the reward signal from the discriminator is not such informative, many methods such as LeakGAN \cite{leakgan}, RankGAN \cite{rankgan}, and
MaliGAN \cite{maligan} were proposed to pass a more informative signal to the generator.
Furthermore, some studies like Gumbel- Softmax GAN \cite{gsgan}, TextGAN \cite{textgan}, FM-GAN \cite{fmgan}, and RelGAN \cite{RelGAN} do not utilize an RL-based approach, but instead, they apply the Gumbel-Softmax trick (that provides a continuous approximation of the discrete multinomial distribution) to pass the gradient to the generator.
Even though both of the above approaches try to alleviate the gradient passing problem and attempt to make the information flow from the discriminator to the generator more effective, they are failed to take full advantage of the discriminator.

In this paper, a new framework for adversarially training generative models of discrete data called DGSAN is proposed.  In this framework, by considering an explicit distribution for the generator (due to the advantage of finite discrete domains) and finding a closed-form relation between the next generator, the current generator, and the current discriminator, the gradient passing problem is resolved.
In the proposed method, the generator and the discriminator are unified in a single network. This network provides the probability distribution of data while also prepares the (conditional) probability of assigning the input to the class of real data. As a result of this integration, the gradient passing issue is bypassed, and the training stability is achieved.%

Below, we describe the main contributions of the proposed method:
\begin{itemize}
	\item
	In the proposed method, as opposed to the existing GAN-based methods for discrete sequence generation, neither the RL approach nor the Gumbel-Softmax trick is required. Instead, by using the closed-form solution of the new generator, the difficulty of training will be resolved. 
	Among the existing methods, MaliGAN \cite{maligan} is the most related work to ours. However, it has significant differences. MaliGAN updates the generator via an RL approach using a gradient estimator that is based on importance sampling. In contrast, the proposed method uses the optimal solution of the discriminator for the current generator directly. MaliGAN uses two different networks for discriminator and generator. However, DGSAN uses a single network for both of them.
	\item
	In standard GANs \cite{gan}, for each generator, the optimal discriminator can be described according to the generator and the real data distribution (even when the generator is not optimal). In the proposed method, by considering domains like discrete ones for which an explicit generative distribution is obtainable, it would be possible to approximate the real data distribution from the discriminator and the current generator. %
	\item
	The proposed method is not limited to the discrete sequence generation. In those domains such as finite discrete ones for which an explicit generative model is considered, the generator can be trained by the proposed adversarial method instead of the maximum likelihood estimation.
\end{itemize}

\section{Related Work} \label{related_work}

As mentioned above, the GANs in discrete domains encounter a problem in propagating the gradient into the generator. SeqGAN \cite{seqgan} has overcome this problem by using a REINFORCE-like algorithm in the training of the generator. It assumes the generator as an agent that receives more reward from the discriminator to generate more realistic sentences and uses the Monte Carlo tree search to estimate the expected reward. %
This method has some difficulties, such as reward sparsity and high variance of training. 
Multiple studies were carried on this approach and tried to improve the RL-based solution by transferring more information from the discriminator to the generator. RankGAN \cite{rankgan} trains a ranker instead of a discriminator which relatively assigns a higher score to the more realistic sequences. In other words, the score of a sentence shows how realistic it is compared to other sentences in the current batch of data. Therefore, the generator will receive 
a more informative gradient. 
LeakGAN \cite{leakgan} takes advantage of the feudal networks by considering the generator as a two-level module containing a manager and a worker. The feature layer of the discriminator is fed to the manager as leaked information. 
Boundary-Seeking GAN \cite{boundary} has also changed the generator's objective using the discriminator's output to make the generator closer to the approximated real data distribution by minimizing $\mathcal{D}_{f}(\widetilde{P} || Q)$ where $Q$ shows the generator. This objective is optimized via an importance sampling approach. There exists a very similar method to Boundry-Seeking GAN called MaliGAN \cite{maligan}, which was mainly proposed for sequence generation. MaliGAN reduces the variance of training by employing various techniques such as Mixed MLE-MALI training and estimating reward with action-value function $Q(s, a)$.
Although this method may show similarities to ours, it still uses an RL-based approach and does not take the complete information from the discriminator. In the proposed method, when the previous generator's distribution is available, the next generator estimating the real data distribution is straightforwardly reachable, and no additional $KL$ optimization is needed. Hence, none of the stabling tricks utilized in these methods, are not required.

Studies like TextGAN~\cite{textgan}, FM-GAN~\cite{fmgan}, and RELGAN~\cite{RelGAN} employ the Gumbel-softmax trick to prepare the gradient signal for the generator. TextGAN \cite{textgan} proposes a new objective for the generator. Its generator tries to make the feature distribution of generated data closer to that of the real data according to the Maximum Mean Discrepancy (MMD) measure. %
TextGAN attempts to push the generator's focus from the last layer of the discriminator to its last feature layer. 
FM-GAN \cite{fmgan} leverages a variant of earth mover's distance (EMD) between the feature distributions of real and synthetic sentences, called feature mover's distance (FMD), as a new generator's objective. RELGAN~\cite{RelGAN} incorporates a relational memory as a new component for modeling the long-distance dependency. Furthermore, its discriminator utilizes multiple representations to prepare a more informative signal for the generator. 

All recent GAN-based methods for the discrete sequence generation, attempt to strengthen the information flow from discriminator to the generator to provide a more informative training signal.
Nonetheless, they were failed to take full advantage of the discriminator. Despite the previous works, in our method, each generator has a hybrid nature. From one perspective, it can be seen as a discriminator, while in another perspective, it is an explicit generative model. Hence, the generator and discriminator are combined in a single model, and all information is pushed to the generator accordingly.

In addition to the GAN-based methods, there are the latent-based generative models, and among them, the Variational Auto-Encoder (VAE) is the most popular one. %
Since the calculation of the likelihood of real samples or the posterior distribution requires the tractability of the partition function, several approaches have been proposed to train these models; e.g. the VAE optimizes the lower bound of the likelihood called \textit{ELBO} \cite{vae}. However, VAE encounters a difficulty in the text generation problem called the \textit{KL Collapse} problem \cite{bowman}. As a consequence of this problem, the decoder ignores the latent space and operates independently. Various studies have been done to identify the main origin of this issue \cite{vae31, vae10, vae11}. %
From these studies, the decoder's universality in conjunction with the undesired global and local optimums has been considered as a source of this difficulty.

\section{Preliminaries on GAN}
In the standard GANs, we have some samples from the real data distribution $P$, which we need to learn. The generator attempts to learn a network $Q$, which can generate samples similar to the real ones, and the discriminator intends to determine whether a sample is real or synthetic. Formally, the objective is \cite{gan}:
\begin{align}\label{eq:cross-ent}
	V(Q, D) = \ME_{x \sim P}[\log D(x)] + \ME_{x \sim Q}[\log (1 - D(x))],
\end{align}
which is maximized with respect to $D$ and minimized with respect to $Q$.
It is shown that the optimal $D$ for a fixed distribution $Q$ is found as \cite{gan}:
\begin{align} \label{eq:optimal_d}
	D^*_Q(x) = \frac{p(x)}{p(x) + q(x)}.
\end{align}
Moreover, for the optimal discriminator $D^*_Q$, minimizing $V(Q,D^*_Q)$ w.r.t. $Q$ leads to the minimization of Jensen-Shannon divergence between $P$ and $Q$ \cite{gan}.
\section{Proposed Method}
In the following subsections, we first describe the general framework for adversarially training explicit generative models.
Then, the theoretical analysis of the proposed training approach is presented. 
Finally, using the proposed framework, a method for sequence generation is presented.

\subsection{Basic Model}
\label{section:alg}
\begin{algorithm}[tb]
   \caption{DGSAN general algorithm}
   \label{alg:overall}
\begin{algorithmic}[1]
    \STATE {\bfseries Input:}
    $B$: Batch size,
    $D$: Number of DGSAN iterations
    
    \STATE Initialize two random generator $Q^{\text{old}}$, and $Q_\theta$.
    \FOR{$D$ iterations}
        \WHILE{not converged}
            \STATE Sample ${\{\mbf{x}_r^{(i)}\}}_{i=1}^B$ from real data.
            \STATE Generate ${\{\mbf{x}_f^{(i)}\}}_{i=1}^B$ samples from $Q^{\text{old}}$.     
            \STATE Train $Q_\theta$ by optimizing Eq.~\ref{eq:obj2}. 
            \\
            Use ${\{\mbf{x}_r^{(i)}\}}_{i=1}^B$ for the first term, and ${\{\mbf{x}_f^{(i)}\}}_{i=1}^B$ for the second term to estimate the expectation.
        \ENDWHILE
        \STATE $Q^{\text{old}} = Q_\theta$
    \ENDFOR
\end{algorithmic}
\end{algorithm}
\begin{algorithm*}[tb]
   \caption{DGSAN sequential data algorithm}
   \label{alg:sequence}
\begin{algorithmic}[1]
   \STATE {\bfseries Input:} 
   $B$: Batch size,
   $M$: Maximum length of the given sequence,
   $D$: Number of DGSAN iterations per length,
   $T$: Sampling temperature

   \STATE Initialize two random generators $Q^{\text{old}}$, and $Q_\theta$.
   
    \STATE $l=1$
    \REPEAT
        \FOR{$D$ iterations}
            \WHILE{not converged}
                \STATE Sample ${\{\mbf{s}^{(i)}\}}_{i=1}^B$ from real data.
                \STATE Select the value of $k$ randomly from $\{0, 1, ..., M-l\}$, as the length of the prefix sequence.
                \STATE Split elements of ${\{\mbf{s}^{(i)}\}}_{i=1}^B$ by considering tokens between $k$ and $k+l$ as $\mbf{x}_r^{(i)} = \mbf{s}^{(i)}_{k:k+l}$, and the first $k$ tokens as $\mbf{c}^{(i)} = \mbf{s}^{(i)}_{1:k}$.
                \STATE Generate ${\{\mbf{x}_f^{(i)}\}}_{i=1}^B$ samples from $Q^{\text{old}}$ conditioned on ${\{\mbf{c}^{(i)}\}}_{i=1}^B$ with the temperature $T$.
                \STATE Train $Q_\theta$ by optimizing Eq.~\ref{eq:seqobj2}
                \\
                Use ${\{\mbf{x}_r^{(i)}\}}_{i=1}^B$ in the first term and ${\{\mbf{x}_f^{(i)}\}}_{i=1}^B$ in the second term to estimate the expectation in Eq.~\ref{eq:seqobj2}. All probabilities are conditioned on $\mbf{c}^{(i)}$).
            \ENDWHILE
            \STATE $Q^{\text{old}} = Q_\theta$
       \ENDFOR
     \STATE $l++$
    \UNTIL{Not reached to max epoch}
\end{algorithmic}
\end{algorithm*}
We intend to learn an explicit generative model $Q$. To this end, we begin from an initial model and improve it iteratively. 
Consider that the generator found in the last step is called $Q^{\text{old}}$, and we want to find a new generator $Q^{\text{new}}$ from it. A discriminator between the real distribution and $Q^{\text{old}}$ is required to form the objective function for optimizing $Q$. 
The discriminator is modeled as:
\begin{align} \label{eq:d}
	D(x) = \frac{q^{\text{new}}(x)}{q^{\text{new}}(x) + q^{\text{old}}(x)}.%
\end{align}
The intuition behind employing the above discriminator is based on the optimal discriminator of GAN.
Indeed, if we want $Q^{\text{new}}$ to be optimal (i.e. equal to the real distribution $P$), the optimal discriminator can be formulated as in Eq.~\ref{eq:d} (according to Eq.~\ref{eq:optimal_d}).
For optimizing the new generator $q_{\theta}$, the following objective is used (according to Eq.~\ref{eq:cross-ent}):
\begin{align*}
		\max_{\theta} \; &\ME_{x \sim P} \Big[\ln{D(x)}\Big]+ \ME_{x \sim Q^{\text{old}}} \Big[\ln{\big(1-D(x)\big)} \Big],
		\numberthis \label{eq:obj2_0}
\end{align*}
which can be reformulated as below (according to Eq.~\ref{eq:d}):
\begin{align*}
		\max_{\theta} \; &\ME_{x \sim P} \Big[\ln{\frac{q_{\theta}(x)}{q_{\theta}(x) + q^{\text{old}}(x)}} \Big] \\
		+ &\ME_{x \sim Q^{\text{old}}} \Big[\ln{\frac{q^{\text{old}}(x)}{q_{\theta}(x) + q^{\text{old}}(x)}} \Big].
		\numberthis \label{eq:obj2}
\end{align*}
Then, the generator $Q_\theta$ is found using samples of the real distribution $P$ and the old generator. 
This procedure is iteratively done. In each DGSAN iteration, the $Q_\theta$ of the last iteration is considered as $Q^{\text{old}}$, and a new $Q_\theta$ is learned.
According to Eq.~\ref{eq:obj2}, since samples come from $Q^{\text{old}}$ and optimization is based on $\theta$, the difficulty of gradient propagation through the discreteness of $Q_\theta$'s output (new generator) is bypassed. %
In the implementation \textcolor{black}{phase}, we use an equal objective based on Softplus \cite{DBLP:conf/nips/DugasBBNG00}:
\begin{align*}
	\min_{\theta}\;  &\ME_{x \sim P}\Big[ \text{Softplus}\big(\ln{q^{\text{old}}(x)} - \ln{q_{\theta}(x)} \big)\Big]\\
	+ &\ME_{x \sim Q^{\text{old}}}\Big[ \text{Softplus}\big(\ln{q_{\theta}(x)} - \ln{q^{\text{old}}(x)} \big)\Big].
	\numberthis \label{eq:mainobj_stable}
\end{align*}

An overview of the training procedure is presented in \mbox{Alg.~\ref{alg:overall}}.
\subsection{Theoretical Analysis}
\label{section:theory}
This section will discuss the convergence of the training algorithm, proposed in Section~\ref{section:alg} (Algorithm~\ref{alg:overall}). The proofs are provided in Appendix.

\begin{theorem}\label{theorem:bregman_js_lower}
	Let $P, Q^{\text{old}},$ and $Q_{\theta}$ denote three distributions. %
	A relation between these distributions can be formulated as:%
	
	\begin{align*}
			&\mathcal{D}_{JS} (P||Q^{\text{old}}) \\ 
			&=\mathcal{L}(P,Q^{\text{old}},Q_{\theta}) + \ME_{x \sim Q^{\text{old}}} \Big[\mathcal{B}_f(\frac{p(x)}{q^{\text{old}}(x)} || \frac{q_{\theta}(x)}{q^{\text{old}}(x)})\Big] ,
			\\
			&= \mathcal{L}(P,Q^{\text{old}},Q_{\theta}) + \ME_{x \sim P} \Big[\mathcal{B}_f(\frac{q^{\text{old}}(x)}{p(x)} || \frac{q^{\text{old}}(x)}{q_{\theta}(x)})\Big],
			\numberthis \label{eq:js}
	\end{align*}
	where $\mathcal{D}_{JS}$ denotes the Jensen–Shannon divergence, $\mathcal{B}_f$ shows the Bregman divergence, and $f(u) = u \ln u - (u+1)\ln(u+1)$ is the function used in the Bregman divergence. Moreover, $\mathcal{L}$ is defined as:
	\begin{align*}
			\mathcal{L}(P,Q^{\text{old}},Q_{\theta}) =
			&\ME_{x \sim P}\big[\ln{\frac{q_{\theta}(x)}{q_{\theta}(x) + q^{\text{old}}(x)}}\big]\\
			+ &\ME_{x \sim Q^{\text{old}}}\big[\ln{\frac{q^{\text{old}}(x)}{q_{\theta}(x) + q^{\text{old}}(x)}}\big].
			\numberthis \label{eq:js:l}
	\end{align*}
\end{theorem}

\begin{corollary}\label{corollary:bregman_js_lower}
 When maximizing $\mathcal{L}(P,Q^{\text{old}},Q_{\theta})$ w.r.t. $\theta$, 
 $\frac{p(x)}{q^{\text{old}}(x)}$ is estimated by $\frac{q_{\theta}(x)}{q^{\text{old}}(x)}$.
\end{corollary}

According to the above corollary, we can use Eq.~\ref{eq:obj2} to find the best new generator for the current generator. Thus, if we reach the global optimum, $Q_{\theta}$ will be the real data distribution. Moreover, based on the latter corollary, adding noise to the distribution $Q^{\text{old}}$ can help the training process in the points where $Q^{\text{old}}$ is near to zero.

Since it is ideal but may be impossible to reach the global optimum of $\mathcal{L}(P,Q^{\text{old}},Q_{\theta})$, the next theorem is provided to support the convergence of the method in the case of reaching a non-optimal solution.
By looking at each iteration from the discriminator training viewpoint, we will have the following theorem.
\begin{theorem}\label{theorem:converg_js}
	If $D_{\theta}(x)=\frac{q_{\theta}(x)}{q_{\theta}(x) + q^{\text{old}}(x)}$ is between a random and an optimal discriminator $D^{*}(x) = \frac{p(x)}{p(x) + q^{\text{old}}(x)}$ for $Q^{\text{old}}$, 
	$\mathcal{D}_{JS}(P||Q_{\theta})$ will be strictly less than $\mathcal{D}_{JS}(P||Q^{\text{old}})$.
\end{theorem}
This theorem shows that when the discriminator is non-optimal, under the above condition, the Jensen-Shannon divergence decreases after each iteration of Algorithm~\ref{alg:overall}, and thus, $ Q $ finally converges to $ P $. %

\subsubsection{The $\mbf{f}$ Family} \label{f_divergence}
We can also extend the loss function and the above theorems to more general ones supporting all strictly convex $f$, instead of $f(u) = u \ln u - (u+1)\ln(u+1)$.
More precisely, inspired by several works on the original GAN \cite{breg-ineq,fgan,improved_ratio_gan,boundary,maligan} that have extended the GAN's initial framework, we also extend the presented framework and theorems in Section~\ref{section:theory} to $f$-divergence family that is more general than the JS divergence $\mathcal{D}_{JS}$.%

\begin{theorem}\label{theorem:bregman_fdiv_lower}
	For every $P, Q^{\text{old}},$ and $Q_{\theta}$ distributions, and $f$-divergence with strictly convex $f$, 
	\begin{align*}
			&\mathcal{D}_f(P \Vert Q^{\text{old}}) =
			\mathcal{L}_f(P, Q^{\text{old}}, \tau_\theta)
			+
			\ME_{x \sim Q^{\text{old}}} \Big[\mathcal{B}_f(
			\frac{p(x)}{q^{\text{old}}(x)}
			\Vert
			\frac{q_{\theta}(x)}{q^{\text{old}}(x)}
			)\Big],
			\\
			&\mathcal{L}_f(P, Q^{\text{old}}, \tau_\theta) = 
			\ME_{x \sim P} \big[\tau_\theta(x)\big]
			- \ME_{x \sim Q^{\text{old}}} \big[f^*(\tau_\theta(x))\big],
			\numberthis \label{eq:theorem:bregman_fdiv_lower}
	\end{align*}
	where $\tau_\theta(x) = f'(\frac{q_{\theta}(x)}{q^{\text{old}}(x)})$, and $f^*$ is the Fenchel conjugate of $f$.
\end{theorem}
It is worth mentioning that \cite{breg-ineq} also presents another variation of Theorem \ref{theorem:bregman_fdiv_lower}.

By maximizing $\mathcal{L}_f(P, Q^{\text{old}}, \tau_\theta)$ respect to the $\theta$, the corresponding Bregman divergence will be minimized and the desired $p(x)$ estimation is achieved.

\begin{theorem}\label{theorem:converg_fdiv}
	If  $D_{\theta}(x)=\frac{q_{\theta}(x)}{q_{\theta}(x) + q^{\text{old}}(x)}$ is between random and optimal discriminator $D^{*}(x) = \frac{p(x)}{p(x) + q^{\text{old}}(x)}$ for $Q^{\text{old}}$, 
	we will have $\mathcal{D}_{f}(P||Q_{\theta}) < \mathcal{D}_{f}(P||Q^{\text{old}})$ for every $f$-divergence by strictly convex $f$.
\end{theorem}

\subsection{Method for sequence generation}
In theory, By considering $\mbf{x}$ as $x_1, x_2, ..., x_M$ which they are tokens of the sequence, we can apply our method on the sequences to learn the joint distribution $P(x_1, x_2, ..., x_M)$.
However, learning such a high dimensional distribution is not feasible. %
Therefore, We train the single token distribution $P(x_i | x_{1:i-1})$ firstly, then the distribution on the sequences of two tokens  $P(x_i, x_{i+1} | x_{1:i-1})$, and so forth (i.e. curriculum learning).

By considering all of the distributions in Eq.~\ref{eq:obj2}, conditioned on a prefix of real data, the generator's objective is obtained as follows:

\begin{align*}
\max_{\theta}\;
\ME_{c \sim P_{prefix}} \Big[
    &\ME_{\mathbf{x} \sim P|c} \big[\ln{\frac{q_{\theta}(\mathbf{x}|c)}{q_{\theta}(\mathbf{x}|c) + q^{\text{old}}(\mathbf{x}|c)}} \big]\\
    + &\ME_{\mathbf{x} \sim Q^{\text{old}}|c} \big[\ln{\frac{q^{\text{old}}(\mathbf{x}|c)}{q_{\theta}(\mathbf{x}|c) + q^{\text{old}}(\mathbf{x}|c)}} \big]
\Big] \numberthis \label{eq:seqobj1}\\
    = &\ME_{\mathbf{x},c \sim P_{prefix}} \big[\ln{\frac{q_{\theta}(\mathbf{x}|c)}{q_{\theta}(\mathbf{x}|c) + q^{\text{old}}(\mathbf{x}|c)}} \big]\\
    + &\ME_{c \sim P_{prefix}  \mathbf{x} \sim Q^{\text{old}}|c} \big[\ln{\frac{q^{\text{old}}(\mathbf{x}|c)}{q_{\theta}(\mathbf{x}|c) + q^{\text{old}}(\mathbf{x}|c)}} \big],
    \numberthis \label{eq:seqobj2}
\end{align*}
where $P_{prefix}$ denotes any prefix of sequences from real data and $\mathbf{x}$ shows conditional samples of length $l$.

In the above objective, we want to learn $q_\theta(x_{k:k+l} | x_{1:k-1})$ distribution for different values of $k$ and $l$. By considering $q_\theta(x_{k:k+l} | x_{1:k-1})=\prod_{i=k}^{k+l}q_\theta(x_i | x_{1:i-1})$, we can rewrite all distributions based on the (conditional) distributions of single tokens.

The condition sequence $x_1,...,x_k$ is first embedded as $\textbf{e}_1,...,\textbf{e}_k$ and then mapped into the sequence of hidden states $\textbf{h}_1,...,\textbf{h}_k$ using a recurrent unit $\textbf{h}_i=g(\textbf{h}_{i-1},\textbf{e}_{i-1})$ (in which $x_0$ shows the start token) and the conditional distribution is modeled as $q_\theta(x_i | x_{1:i-1})=softmax(\textbf{V}\textbf{h}_i)$. Let $\theta$ denote the set of parameters of the generator network (containing the parameters of $g$ and the matrix $\textbf{V}$).%

Usually sequence modeling is converted to learning $q(x_k | x_{1:k-1})$ whereas the joint distribution can be rewritten as $q(x_{1:M})=\prod_{k=1}^{M}q(x_k | x_{1:k-1})$.
Since the conditioned sequence always brings from real data during training while it is just the generated sequence by the model in the test time, there is a well-known problem %
called exposure bias. In our method, we incrementally raise the length $l$ of the predicting sequence since increasing this length alleviates the exposure bias problem consequently. Increasing the predicting sequence's length leads the model to see generated samples during the training phase. In the ideal case, if the length of the predicting sequence is equal to the target sequence length, exposure bias vanish entirely.

An overview of our training procedure for generating discrete sequences is provided in \mbox{Algorithm~\ref{alg:sequence}}.

\section{Experiments and Results}
\begin{table*}[!htb]
\centering
\caption{Performance of models (using different measures) on \textit{Amazon} dataset.}\label{table:Amazon}
\small\tabcolsep=0.07cm
\begin{tabular}{||c||c|c|c c c|c c c|c c c||}\hline\hline Method	& NLL	& FBD	& MSJ3	& MSJ5	& MSJ7	& BL3	& BL5	& BL7	& BBL3	& BBL5	& BBL7\\
\hline\hline
DGSAN	& $113.146$	& $\mathbf{0.420}$	& $0.442$	& $0.254$	& $\mathbf{0.138}$	& $\mathbf{0.945}$	& $\mathbf{0.728}$	& $\mathbf{0.442}$	& $0.724$	& $0.416$	& $\mathbf{0.209}$ \\
\hline
MLE	& $\mathbf{99.369}$	& $3.254$	& $\mathbf{0.560}$	& $\mathbf{0.265}$	& $0.121$	& $0.768$	& $0.397$	& $0.174$	& $0.786$	& $\mathbf{0.425}$	& $0.193$ \\
\hline
VAE	& -	& $2.579$	& $0.556$	& $0.264$	& $0.121$	& $0.773$	& $0.399$	& $0.177$	& $\mathbf{0.787}$	& $\mathbf{0.425}$	& $0.192$ \\
\hline
MaliGAN	& $123.069$	& $8.951$	& $0.313$	& $0.147$	& $0.069$	& $0.740$	& $0.407$	& $0.188$	& $0.686$	& $0.311$	& $0.136$ \\
\hline
RankGAN	& $127.364$	& $6.511$	& $0.307$	& $0.141$	& $0.064$	& $0.773$	& $0.363$	& $0.158$	& $0.669$	& $0.303$	& $0.129$ \\
\hline
SeqGAN	& $141.881$	& $9.668$	& $0.294$	& $0.140$	& $0.064$	& $0.807$	& $0.419$	& $0.177$	& $0.681$	& $0.324$	& $0.141$ \\
\hline
\hline\end{tabular}\normalsize 
 \end{table*}

\begin{table*}[!htb]
\centering
\caption{Performance of models (using different measures) on \textit{Yelp Restaurant} dataset.}\label{table:Yelp Restaurant}
\small\tabcolsep=0.07cm
\begin{tabular}{||c||c|c|c c c|c c c|c c c||}\hline\hline Method	& NLL	& FBD	& MSJ3	& MSJ5	& MSJ7	& BL3	& BL5	& BL7	& BBL3	& BBL5	& BBL7\\
\hline\hline
DGSAN	& $56.398$	& $\mathbf{0.382}$	& $0.334$	& $\mathbf{0.152}$	& $\mathbf{0.067}$	& $\mathbf{0.756}$	& $\mathbf{0.417}$	& $\mathbf{0.245}$	& $0.538$	& $\mathbf{0.259}$	& $\mathbf{0.157}$ \\
\hline
MLE	& $\mathbf{50.201}$	& $2.276$	& $0.322$	& $0.125$	& $0.045$	& $0.542$	& $0.246$	& $0.149$	& $0.551$	& $0.245$	& $0.147$ \\
\hline
VAE	& -	& $2.297$	& $\mathbf{0.344}$	& $0.133$	& $0.048$	& $0.554$	& $0.242$	& $0.145$	& $\mathbf{0.559}$	& $0.251$	& $0.150$ \\
\hline
MaliGAN	& $55.150$	& $1.351$	& $0.226$	& $0.086$	& $0.031$	& $0.562$	& $0.291$	& $0.191$	& $0.484$	& $0.210$	& $0.130$ \\
\hline
RankGAN	& $58.463$	& $1.804$	& $0.250$	& $0.101$	& $0.035$	& $0.678$	& $0.333$	& $0.206$	& $0.466$	& $0.209$	& $0.129$ \\
\hline
SeqGAN	& $53.135$	& $1.022$	& $0.276$	& $0.114$	& $0.042$	& $0.617$	& $0.307$	& $0.198$	& $0.504$	& $0.226$	& $0.138$ \\
\hline
\hline\end{tabular}\normalsize 
 \end{table*}

\begin{table*}[!htb]
\centering
\caption{Performance of models (using different measures) on \textit{Coco} dataset.}\label{table:Coco}
\small\tabcolsep=0.07cm
\begin{tabular}{||c||c|c|c c c|c c c|c c c||}\hline\hline Method	& NLL	& FBD	& MSJ3	& MSJ5	& MSJ7	& BL3	& BL5	& BL7	& BBL3	& BBL5	& BBL7\\
\hline\hline
DGSAN	& $\mathbf{73.630}$	& $\mathbf{1.949}$	& $0.208$	& $0.091$	& $\mathbf{0.035}$	& $\mathbf{0.647}$	& $\mathbf{0.307}$	& $\mathbf{0.154}$	& $0.489$	& $0.206$	& $0.107$ \\
\hline
MLE	& $99.525$	& $3.209$	& $0.224$	& $0.088$	& $0.029$	& $0.517$	& $0.209$	& $0.105$	& $0.549$	& $0.219$	& $0.109$ \\
\hline
VAE	& -	& $2.559$	& $\mathbf{0.230}$	& $\mathbf{0.094}$	& $0.033$	& $0.530$	& $0.215$	& $0.109$	& $\mathbf{0.568}$	& $\mathbf{0.234}$	& $\mathbf{0.116}$ \\
\hline
MaliGAN	& $93.243$	& $3.931$	& $0.176$	& $0.065$	& $0.021$	& $0.503$	& $0.200$	& $0.100$	& $0.504$	& $0.192$	& $0.098$ \\
\hline
RankGAN	& $103.071$	& $3.198$	& $0.118$	& $0.038$	& $0.010$	& $0.474$	& $0.186$	& $0.098$	& $0.381$	& $0.135$	& $0.074$ \\
\hline
SeqGAN	& $93.681$	& $2.344$	& $0.175$	& $0.067$	& $0.021$	& $0.524$	& $0.229$	& $0.120$	& $0.440$	& $0.169$	& $0.089$ \\
\hline
\hline\end{tabular}\normalsize 
 \end{table*}
In this section, we conduct experiments to examine the proposed method for the sequence generation problem and compare it to the state-of-the-art methods. 

\subsection{Evaluation Metrics} \label{measures}
First of all, Negative Log-Likelihood (NLL) as a popular metric for evaluating generative models is presented, then three n-gram based measures for evaluating sequence generation are introduced. Finally, a metric that considers the semantic is described.
\subsubsection{Negative Log Likelihood (NLL)}
It shows the negative log-likelihood of real data in the model. The GAN-based methods tend to have poor NLL scores on the training/test data since this measure is much sensitive to the mode collapse phenomenon~\cite{GANFalling}. It should be noted that the NLL is closely related to another well-known metric called Perplexity (so, we just report NLL to avoid redundancy).

\subsubsection{BLEU (BL)}
BLEU is a metric designed to evaluate the machine translation models \cite{bleu}. \textcolor{black}{It measures n-gram similarities of a test sentence to a reference set and then takes the geometric mean of n-gram similarities to produce a score for the test sentence}. As discussed in \cite{GANFalling}, a model repeatedly generating one high-quality sentence can get a high BLEU score while completely losing diversity. Therefore, BLEU evaluates the quality of the samples and is not sensitive to their diversity \cite{texygen, bert_score, joint_measure}.%

\subsubsection{Backward BLEU (BBL)}
The Backward BLEU was introduced by \cite{Shi_2018} to measure the diversity of generated samples. It considers generated samples as the reference and evaluates each test sample by BLEU.

\subsubsection{MS-Jaccard (MSJ)}
Either having a model, repeatedly generating just one high-quality sample or having a model generating a wide variety of low-quality samples, is disappointing. Thus, to consider this trade-off and have a metric jointly measuring the validity and coverage, the recently proposed MS-Jaccard metric is taken into account \cite{joint_measure}. The n-grams of generated samples and real samples are considered as two multi-sets which preserve the repetition of n-grams. 
Then, the similarity of the resulted multi-sets is computed by the Jaccard similarity measure whose higher scores are more desired.
If the generated sentences do not have diversity (e.g., when the mode collapse happens) or lose quality, the n-gram distribution of generated texts will be different from that of the real texts and this measure will be decreased. %

\subsubsection{Fr\'echet BERT Distance (FBD)}
FBD \cite{joint_measure} is inspired by FID \cite{heusel2017gans} which is a famous metric in image generation task.
Unlike other metrics, the FBD considers the semantic of sentences.
The FBD considers the distance between the distributions of real and generated text in a feature space found by BERT (as a powerful language representation model) \cite{devlin2018bert}. Indeed, the distance is the Fr\'echet distance between the two Gaussian distributions fitted to real and generated data in the feature space.

\subsection{Datasets}
We have used three real-world datasets: Image COCO captions\footnote{\url{https://github.com/pclucas14/GansFallingShort}}, Yelp restaurant reviews\footnote{\url{https://github.com/shentianxiao/language-style-transfer}}\cite{ShenLBJ17}, and Amazon Review\footnote{\url{https://github.com/williamSYSU/TextGAN-PyTorch}} \cite{McAuleyL13} to cover a wide range of linguistic datasets. 
Table \ref{datasets_statics} shows the statistics of datasets.
\begin{table}[!htbp]
\begin{tabular}{|c|c|c|c|}
\hline
\textbf{Dataset} &
  \begin{tabular}[c]{@{}c@{}}Amazon\\ Review\end{tabular} &
  \begin{tabular}[c]{@{}c@{}}Image\\ COCO\\ captions\end{tabular} &
  \begin{tabular}[c]{@{}c@{}}Yelp\\ restaurant\\ reviews\end{tabular} \\ \hline
\textbf{Train set size}      & 174588 & 10000 & 40000 \\ \hline
\textbf{Validation set size} & 19436  & 10000 & 10000 \\ \hline
\textbf{Test set size}       & 19436  & 10000 & 10000 \\ \hline
\textbf{Vocabulary size}     & 6282   & 6488  & 7802  \\ \hline
\textbf{Max Length}          & 41     & 38    & 30    \\ \hline
\textbf{Mean Length}         & 28.87  & 11.62 & 9.45  \\ \hline
\end{tabular}\caption{Datasets Statistics}\label{datasets_statics}
\end{table}

\subsection{Experiment Setup}
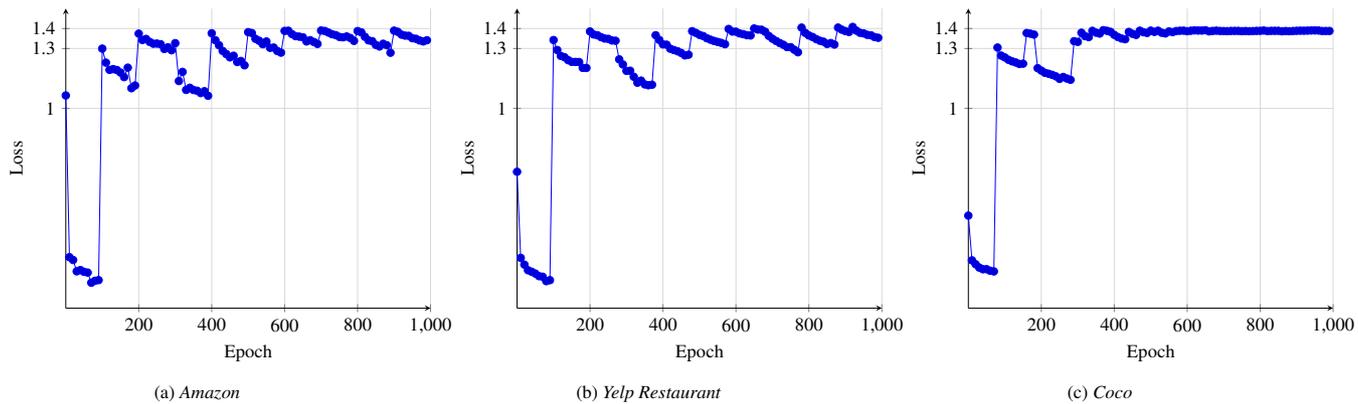
\begin{figure*}[!ht]
    \centering
    \begin{subfigure}[t]{0.3\textwidth}\centering
    \begin{tikzpicture}[scale=0.7]
    \begin{axis}[grid=both,
                 grid style={solid,gray!30!white},
                 axis lines=middle,
                 xlabel={Epoch},
                 ylabel={Loss},
                 x label style={at={(axis description cs:0.5,-0.1)},anchor=north},
                 y label style={at={(axis description cs:-0.1,.5)},rotate=90,anchor=south},
                 ymin=0,
                 ymax=1.5,
                 ytick={0,1,1.3,1.4},
                 xmin=0,
                 xmax=1000,
                ]
      \addplot table [x=Step, y=Value, col sep=comma] {loss/coco.csv};
    \end{axis}
    \end{tikzpicture}
    \caption{\textit{Amazon}}
    \end{subfigure}
    \hfill
    \begin{subfigure}[t]{0.3\textwidth}\centering
    \begin{tikzpicture}[scale=0.7]
    \begin{axis}[grid=both,
                 grid style={solid,gray!30!white},
                 axis lines=middle,
                 xlabel={Epoch},
                 ylabel={Loss},
                 x label style={at={(axis description cs:0.5,-0.1)},anchor=north},
                 y label style={at={(axis description cs:-0.1,.5)},rotate=90,anchor=south},
                 ymin=0,
                 ymax=1.5,
                 ytick={0,1,1.3,1.4},
                 xmin=0,
                 xmax=1000,
                ]
      \addplot table [x=Step, y=Value, col sep=comma] {loss/yelp_restaurant.csv};
    \end{axis}
    \end{tikzpicture}
    \caption{\textit{Yelp Restaurant}}
    \end{subfigure}
    \hfill
    \begin{subfigure}[t]{0.3\textwidth}\centering
    \begin{tikzpicture}[scale=0.7]
    \begin{axis}[grid=both,
                 grid style={solid,gray!30!white},
                 axis lines=middle,
                 xlabel={Epoch},
                 ylabel={Loss},
                 x label style={at={(axis description cs:0.5,-0.1)},anchor=north},
                 y label style={at={(axis description cs:-0.1,.5)},rotate=90,anchor=south},
                 ymin=0,
                 ymax=1.5,
                 ytick={0,1,1.3,1.4},
                 xmin=0,
                 xmax=1000,
                ]
      \addplot table [x=Step, y=Value, col sep=comma] {loss/amazon_app_book.csv};
    \end{axis}
    \end{tikzpicture}
    \caption{\textit{Coco}}
    \end{subfigure}
    \caption{DGSAN training loss of different datasets}
    \label{losses}
\end{figure*}
\label{experiment_setup}
Several experiments are conducted to compare our method with some popular GAN-based methods, MLE, and VAE for sequence generation.

The generator networks (i.e. the decoder in VAE and the generator in MLE, GANs, and DGSAN) are the same and implemented with one-layer LSTM, having 128 embedding and 64 hidden dimension.

All GANs including SeqGAN, MaliGAN, and RankGAN are implemented in Texygen framework\footnote{\url{https://github.com/geek-ai/Texygen}}\cite{texygen}
and have a discriminator with two CNN layers including the filter size $2$ and $3$ and the number of filters $100$ and $200$, respectively.

The VAE's encoder is an LSTM with 128 embedding and 128 hidden dimension, and the latent dimension is 64.
To mitigate the \textit{KL Collapse} problem, it is trained with \textit{KL term anealing} trick which is implemented with \textit{Tanh} function. Also, word dropout is taken into account.
Its implementation is based on \href{https://github.com/kefirski/pytorch_RVAE}{this  Github repository\footnote{\url{https://github.com/kefirski/pytorch_RVAE}}}.

All evaluations are based on the best checkpoints of models concerning the FBD score which is evaluated every 5 epochs on validation data during training. In our experiments, the number of DGSAN iterations
per length ($ D $) was set to 5, and the temperature in DGSAN training ($ T $) was chosen as $2.0$.

MLE, VAE, and DGSAN are trained for 1000 epochs on all datasets. Since the GANs include the discriminator training phase, each epoch of their training lasts much more. Therefore, to have a fair training time for all models, fewer epochs are considered for GANs. They are trained for 280 epochs including 80 epochs of the MLE pre-training. Moreover, training of the GANs usually ends in the mode collapsed state, and the best FBD based checkpoint is not near the end. This is approved with our experiments; e.g., the RankGAN achieves the best FBD score around the 115th epoch which is far from the last epoch.

The baseline and evaluation codes is available at \href{https://github.com/Danial-Alh/TextGenerationEvaluator}{this Github repository\footnote{\url{https://github.com/Danial-Alh/TextGenerationEvaluator}}}. 
Also, the implementation code of DGSAN is available in \href{https://github.com/IAmS4n/DGSAN}{this  Github repository\footnote{\url{https://github.com/IAmS4n/DGSAN}}}.

\subsection{Results}
As discussed in Section \ref{experiment_setup}, all models are selected based on the best FBD score, since it not only evaluates the semantic and syntax validity of samples, but also evaluates the diversity of them simultaneously \cite{joint_measure}. Moreover, as the FBD considers the whole sentence, it is independent of any length parameter selection, such as $n$ in the N-gram metrics.

Tables \ref{table:Amazon}, \ref{table:Yelp Restaurant}, and \ref{table:Coco} show the results of different methods on the Amazon, Yelp Restaurant, and the Coco datasets, respectively. In addition to the FBD, three N-gram based metrics, including MS-Jaccard (MSJ), BLEU (BL), and Backward BLEU (BBL), are reported with different $n$ of $3, 5, 7$. The number of generated samples is same as the test set size.

Recent studies \cite{GANFalling, joint_measure} show that the regular training of GANs typically leads them to collapse in generating high-quality samples while other baselines such as MLE will perform better when considering both quality and diversity. Although, our experiments are based on the best FBD checkpoints to alleviate this problem for GAN-based methods; nevertheless, their results are not competitive, yet; even with the MLE.
It is noticeable that the DGSAN method has outperformed other methods according to the FBD and BLEU on all of the datasets.
There is also a correlation between the improvement of the DGSAN results and $n$ in the N-gram based metrics. Moreover, as the FBD considers the whole of the sentences, the improvement of DGSAN in sequence generation with longer coherency can be inferred from the results.

Figure \ref{losses} shows the loss values of DGSAN during the training. The loss value approximately converged to the value of $1.38$ which is consistent with the theory (indicating that the discriminator loss when it can not distinguish between real and fake, is $2\ln2 \approx 1.38$). It is worth mentioning that difficulty in reaching to this equilibrium is usually a big challenge for GAN-based methods specially on discrete data. 

Finally, some samples of generated sentences by each of the models are provided in the Appendix. For each of the models and datasets, the provided samples are selected in two ways, the random selection and high-quality selection based on BLEU5 among generated samples.

\section{Conclusion}
In this paper, we proposed a generative adversarial model for domains in which we can consider an explicit distribution and sample from that distribution, including discrete domains. As opposed to the existing approaches, we do not need an RL-based approach for the generator's training.
As a consequence of finding the generator via a closed-form solution in each iteration, we removed the gradient passing to the generator. Moreover, the GAN stability issues during training caused by seeking the saddle point of the objective are bypassed in the proposed method. 

Since generating a sequence of discrete data is an essential task in natural language generation, we examined the proposed method in generating natural language texts. 
Experiments demonstrate that the proposed method can model a more similar distribution to the real distribution (than those generated by the compared methods) according to the measures that approximate the similarity of the distributions.

\bibliography{refs.bib}
\newpage
\section{Appendix}%
Proofs of the theorems presented in Section~\ref{section:theory} are described here.

\begin{lemma}\label{lemma:fconjugate_base_derivation}
	If $f$ is convex then
	\begin{align}
    	f^*(f'(x)) = f'(x)x - f(x),
	\end{align}
	where $f^*$ is the Fenchel conjugate of $f$.
\end{lemma}

\begin{proof}
	For each convex function $f$:
	\begin{align*}
    	f(u) &\geq f^\prime(x) (u-x) + f(x)\\
    	\Rightarrow x f^\prime(x) - f(x) &\geq u f^\prime(x) - f(u)\\
    	\Rightarrow x f^\prime(x) - f(x) &\geq \sup_{u \in dom_f} \{ u f^\prime(x) - f(u)\}. \numberthis
	\end{align*}
	On the other side, it is obvious that for every $x$, we have:
	\begin{align}
    	\sup_{u \in dom_f} \{ u f^\prime(x) - f(u)\} \geq x f^\prime(x) - f(x).
	\end{align}
	Therefore:
	\begin{align}
    	\sup_{u \in dom_f} \{ u f^\prime(x) - f(u)\} = x f^\prime(x) - f(x).
	\end{align}
	From the Fenchel conjugate definition, we obtain:
	\begin{align*}
    	f^*(t) &= \sup_{u \in dom_f} \{ ut - f(u) \}\\
    	\Rightarrow f^*(f^\prime(x)) &= \sup_{u \in dom_f} \{ u f^\prime(x) - f(u) \}\\
    	\Rightarrow f^*(f^\prime(x)) &=  x f^\prime(x) - f(x).
    	\numberthis
	\end{align*}
\end{proof}

\subsection*{\textbf{Proof of Theorem \ref{theorem:bregman_fdiv_lower}}}
\begin{proof}
	According to the Bregman divergence definition, we reach the following equation:
	\begin{align*}
    	\mathcal{B}_f(\frac{p(x)}{q^{\text{old}}(x)} &\Vert \frac{q_\theta(x)}{q^{\text{old}}(x)} ) = \\
    	&f(\frac{p(x)}{q^{\text{old}}(x)}) - f(\frac{q_\theta(x)}{q^{\text{old}}(x)})\\
    	&- f^\prime(\frac{q_\theta(x)}{q^{\text{old}}(x)}) \Big(\frac{p(x)}{q^{\text{old}}(x)}-\frac{q_\theta(x)}{q^{\text{old}}(x)}\Big)\\
    	& = f(\frac{p(x)}{q^{\text{old}}(x)})
    	- \frac{p(x)}{q^{\text{old}}(x)} f^\prime(\frac{q_\theta(x)}{q^{\text{old}}(x)})\\
    	&+ \Big( f^\prime(\frac{q_\theta(x)}{q^{\text{old}}(x)}) \frac{q_\theta(x)}{q^{\text{old}}(x)} - f(\frac{q_\theta(x)}{q^{\text{old}}(x)}) \Big).
    	\numberthis
	\end{align*}
	By using Lemma \ref{lemma:fconjugate_base_derivation}, and defining %
	$\tau_\theta=\frac{q_\theta(x)}{q^{\text{old}}(x)}$, the above Bregman divergence can be reformulated as:
	\begin{align*}\label{eq:bregman_eqaulity_without_expected}
    	&\mathcal{B}_f(\frac{p(x)}{q^{\text{old}}(x)} \Vert \frac{q_\theta(x)}{q^{\text{old}}(x)}) = \\
    	&f(\frac{p(x)}{q^{\text{old}}(x)}) 
    	- \frac{p(x)}{q^{\text{old}}(x)} f^\prime(\frac{q_\theta(x)}{q^{\text{old}}(x)})
    	+ f^*( f^\prime (\frac{q_\theta(x)}{q^{\text{old}}(x)})) \\
    	& = f(\frac{p(x)}{q^{\text{old}}(x)})
    	- \frac{p(x)}{q^{\text{old}}(x)} \tau_\theta(x)
    	+ f^*(\tau_\theta(x)). \numberthis
	\end{align*}
	Hence, by getting expectation, we have:
	\begin{align*}
    	&\ME_{x \sim Q^{\text{old}}} [\mathcal{B}_f(\frac{p(x)}{q^{\text{old}}(x)} \Vert r_\theta(x))]
    	= \ME_{x \sim Q^{\text{old}}} [f(\frac{p(x)}{q^{\text{old}}(x)})]\\
    	&- \ME_{x \sim Q^{\text{old}}} [\frac{p(x)}{q^{\text{old}}(x)} \tau_\theta(x)]
    	+\ME_{x \sim Q^{\text{old}}} [f^*(\tau_\theta(x))] \\
    	&= \ME_{x \sim Q^{\text{old}}} [f(\frac{p(x)}{q^{\text{old}}(x)})]
    	- \ME_{x \sim P} [\tau_\theta(x)]
    	+\ME_{x \sim Q^{\text{old}}} [f^*(\tau_\theta(x))] \\
    	&=
    	\mathcal{D}_f(P \Vert Q^{\text{old}})
    	- \mathcal{L}_f(P, Q^{\text{old}},\tau_\theta).
    	\numberthis
	\end{align*}
\end{proof}

\begin{lemma}\label{lemma:Bregman_inverse_symmetric_f}
	Let $\alpha$ be a constant,
	if $f$ can be written as $f(u) =u f(\frac{1}{u}) + \alpha(u-1)$, then:
	\begin{align}
	    \frac{1}{x}\mathcal{B}_f(x \Vert y)= \mathcal{B}_f(\frac{1}{x} \Vert \frac{1}{y}).
	\end{align}
\end{lemma}

\begin{proof}
	By taking the derivative of $f$:
	\begin{align*}
    	f(u) &= u f(\frac{1}{u}) + \alpha(u-1) \\
    	\Rightarrow f^\prime(u) &= f(\frac{1}{u}) - \frac{f^\prime(\frac{1}{u})}{u} + \alpha,
    	\numberthis
	\end{align*}
	and using the definition of Bregman divergence, we have:
	\begin{align*}
    	\mathcal{B}_f(\frac{1}{x} \Vert \frac{1}{y})
    	&= f(\frac{1}{x}) - f(\frac{1}{y}) - f^\prime(\frac{1}{y}) (\frac{1}{x} - \frac{1}{y}) \\
    	&= \Big(\frac{f(x)}{x} + \frac{\alpha}{x} -\alpha\Big)
    	- \Big(\frac{f(y)}{y} + \frac{\alpha}{y} -\alpha \Big)\\
    	&- \Big( (f(y)-y f^\prime(y)+\alpha)(\frac{1}{x}-\frac{1}{y}) \Big)\\
    	&= \Big(\frac{f(x)}{x} + \frac{\alpha}{x}
    	- \frac{f(y)}{y} - \frac{\alpha}{y}\Big)\\
    	&- \Big(
    	\frac{f(y)}{x} - \frac{y f^\prime(y)}{x} + \frac{\alpha}{x}
    	-\frac{f(y)}{y} + f^\prime(y) - \frac{\alpha}{y}
    	\Big)\\
    	&= \frac{f(x)}{x} - \frac{f(y)}{x} + \frac{y f^\prime(y)}{x} - f^\prime(y) \\
    	&= 
    	\frac{1}{x} \Big(
    	f(x) - f(y) - f^\prime(y) (x-y)
    	\Big) \\
    	&= 
    	\frac{1}{x} \mathcal{B}_f(x \Vert y).
    	\numberthis
	\end{align*}
\end{proof}

\begin{lemma}\label{theorem:symtric_bregman}
	Let $\alpha$ be a constant, if $f$ satisfies $f(u) =u f(\frac{1}{u}) + \alpha(u-1)$, then:
	
	\begin{align}\label{equation:excpected_Bregman_inverse_symmetric_f}
    	\ME_{x \sim Q} [\mathcal{B}_f(\frac{p(x)}{q(x)} \Vert r(x))]
    	=
    	\ME_{x \sim P} [\mathcal{B}_f(\frac{q(x)}{p(x)} \Vert \frac{1}{r(x)})].
	\end{align}
\end{lemma}

\begin{proof}
	Based on Lemma \ref{lemma:Bregman_inverse_symmetric_f}, for each $x$, we have:
	\begin{align}\label{eq:withoutexcpected_Bregman_inverse_symmetric_f}
    	q(x) \mathcal{B}_f(\frac{p(x)}{q(x)} \Vert r(x))
    	=
    	p(x) \mathcal{B}_f(\frac{q(x)}{p(x)} \Vert \frac{1}{r(x)}).
	\end{align}
		Finally, using a summation on all $x$ in Eq. \ref{eq:withoutexcpected_Bregman_inverse_symmetric_f}, the proof is completed.
\end{proof}

\subsection*{\textbf{Proof of Theorem \ref{theorem:bregman_js_lower}}}
\begin{proof}
	Let $f(u) = u \ln u - (u+1) \ln(u+1)$. The Fenchel conjugate of $f$ is obtained as below:
	\begin{align}
    	f^*(t) =- \ln(1-\exp(t)) \quad s.t. \; t\in\mathbb{R}_{-}.
	\end{align}
	Based on Theorem \ref{theorem:bregman_fdiv_lower}, we have:
	\begin{align*}
	    \tau_\theta(x) &= \ln \frac{q_\theta(x)}{q_\theta(x) + q^{\text{old}}(x)}\\
	    \Rightarrow
	    f^*(\tau_\theta(x)) &= - \ln (\frac{q^{\text{old}}(x)}{q_\theta(x) + q^{\text{old}}(x)}),
	    \numberthis
	\end{align*}
	and thus:
	\begin{align*}
    	\mathcal{D}_{JS}(P \Vert Q^{\text{old}}) &=
    	\mathcal{L}_f(P, Q^{\text{old}}, \tau_\theta)\\
    	&+ \ME_{x \sim Q^{\text{old}}} [\mathcal{B}_f(\frac{p(x)}{q^{\text{old}}(x)} \Vert \frac{q_\theta(x)}{q^{\text{old}}(x)})],\\
    	\mathcal{L}_f(P, Q^{\text{old}}, \tau_\theta) &= \ME_{x \sim P} [ \ln  \frac{q_\theta(x)}{q_\theta(x)+q^{\text{old}}(x)} ] \\
    	&+ \ME_{x \sim Q^{\text{old}}} [\ln \frac{q^{\text{old}}(x)}{q_\theta(x)+q^{\text{old}}(x)}],\numberthis
	\end{align*}
	where $\mathcal{D}_{JS}$ is Jensen-Shannon divergence.

	On other side, $f$ satisfies the assumptions of Lemma
	\ref{theorem:symtric_bregman}
	by selecting $\alpha=0$, Thus:
	\begin{align}
    	\ME_{x \sim Q^{\text{old}}} [\mathcal{B}_f(\frac{p(x)}{q^{\text{old}}(x)} \Vert \frac{q_\theta(x)}{q^{\text{old}}(x)} )]
    	=
    	\ME_{x \sim P} [\mathcal{B}_f(\frac{q^{\text{old}}(x)}{p(x)} \Vert \frac{q^{\text{old}}(x)}{q_\theta(x)} )],
	\end{align}
	and proof will be complete.
\end{proof}

\subsection*{\textbf{Proof of Corollary \ref{corollary:bregman_js_lower}}}
\begin{proof}
	Since $\mathcal{D}_{JS} (P||Q^{\text{old}})$ is constant with respect to $\theta$, maximizing $\mathcal{L}(P,Q^{\text{old}},Q_\theta)$ is equivalent to minimizing $\ME_{x \sim P}\mathcal{B}_f(\frac{q^{\text{old}}(x)}{p(x)} || \frac{q^{\text{old}}(x)}{q_{\theta}(x)})$ and $\ME_{x \sim Q^{\text{old}}}\mathcal{B}_f(\frac{p(x)}{q^{\text{old}}(x)} || \frac{q_{\theta}(x)}{q^{\text{old}}(x)})$
	(according to Eq.~\ref{eq:js}).
\end{proof}

\begin{lemma}\label{lemma:good_fdivergence}
	For each $f$-divergence with function $f$, there exists another $f$-divergence with function $g$ defined as below where $\mathcal{D}_f(P \Vert Q) = \mathcal{D}_g(P \Vert Q)$ and $g^\prime(1)=0$.
	\begin{align}
    	g(u) = f^\prime(1) - f^\prime(1) u + f(u).
	\end{align}
\end{lemma}

\begin{proof}
	\begin{align*}
    	\mathcal{D}_{g}(P \Vert Q)
    	&= \sum_x q(x) g(\frac{p(x)}{q(x)}) \\
    	&= \sum_x q(x) [f^\prime(1) - f^\prime(1) \frac{p(x)}{q(x)} + f(\frac{p(x)}{q(x)})]\\
    	&= f^\prime(1) - f^\prime(1) + \sum_x q(x) [f(\frac{p(x)}{q(x)})]\\
    	&= f^\prime(1) - f^\prime(1) + \mathcal{D}_f(P \Vert Q)\\
    	&= \mathcal{D}_f(P \Vert Q). \numberthis
	\end{align*}
	As $f(1) =0$ for a valid $f$-divergence, we will have:
	\begin{align}
    	g^\prime(1) = f^\prime(1) - f^\prime(1) + f(1) =  f(1) =0 .
	\end{align}
\end{proof}

\begin{lemma}\label{lemma:between_divergence_anyf}
	Let $P,Q$ be two distributions; if there exists a distribution $\phi(x)$ that ${\compmin{p(x),q(x)}<\phi(x)<\compmax{p(x),q(x)}}$ for all $x$, we will have:
	\begin{align}
    	\mathcal{D}_{f}(P \Vert \Phi) < \mathcal{D}_{f}(P \Vert Q),
	\end{align}
	where $f$ is strictly convex.
\end{lemma}
\begin{proof}
	by using $f(1)=0$ property:
	\begin{align*}
    	\mathcal{D}_{f}(P \Vert \Phi) - \mathcal{D}_{f}(P \Vert Q) &=\\
    	\sum_x &p(x) [f(\frac{p(x)}{\phi(x)}) - f(\frac{p(x)}{q(x)})]\\
    	=\sum_{x:p(x)>q(x)} &p(x) [f(\frac{p(x)}{\phi(x)}) - f(\frac{p(x)}{q(x)})]\\
    	+\sum_{x:p(x)<q(x)} &p(x) [f(\frac{p(x)}{\phi(x)}) - f(\frac{p(x)}{q(x)})]
    	\numberthis
	\end{align*}
and by using lemma assumption \big(${\compmin{p(x),q(x)}<\phi(x)<\compmax{p(x),q(x)}}$\big),
	\begin{align*}
    	\mathcal{D}_{f}(P \Vert \Phi) - \mathcal{D}_{f}(P \Vert Q)
    	&=\\
    	\sum_{x:p(x)>\phi(x)>q(x)} &p(x) [f(\frac{p(x)}{\phi(x)}) - f(\frac{p(x)}{q(x)})]\\
    	+\sum_{x:p(x)<\phi(x)<q(x)} &p(x) [f(\frac{p(x)}{\phi(x)}) - f(\frac{p(x)}{q(x)})].
    	\label{eq:convergence_partition}\numberthis
	\end{align*}
	As shown in Lemma \ref{lemma:good_fdivergence}, we can consider an equivalent divergence having the property $f^\prime(1)=0$.
	Thus, this lemma and the strict convexity of $f$ conclude that $f$ is strictly increasing function for $x>1$ and strictly decreasing one for $x<1$. As a result, each summation term on Eq.~\ref{eq:convergence_partition}
	has a negative value and the proof is completed.
\end{proof}

\subsection*{\textbf{Proof of Theorem \ref{theorem:converg_fdiv}}}
\begin{proof}
	The function $f(x)=\frac{x}{1-x}$ is strictly increasing between zero and one, so we have:
	\begin{align*}
    	&\compmin{0.5, D^*(x)} < D_\theta(x) < \compmax{0.5, D^*(x)}\\
    	\Rightarrow
    	&\frac{\compmin{0.5, D^*(x)}}{1 - \compmin{0.5, D^*(x)}} < \frac{D_\theta(x)}{1-D_\theta(x)}\\
    	&< \frac{\compmax{0.5, D^*(x)}}{1 - \compmax{0.5, D^*(x)}}\\
    	\Rightarrow
    	&\compmin{\frac{0.5}{1-0.5}, \frac{D^*(x)}{1- D^*(x)}} < \frac{D_\theta(x)}{1-D_\theta(x)} \\
    	&<\compmax{\frac{0.5}{1-0.5}, \frac{D^*(x)}{1- D^*(x)}}\\
    	\Rightarrow
    	&\compmin{1, \frac{p(x)}{q^{\text{old}}(x)}} < \frac{D_\theta(x)}{1-D_\theta(x)}\\
    	&<\compmax{1, \frac{p(x)}{q^{\text{old}}(x)}}\\
    	\Rightarrow
    	&\compmin{q^{\text{old}}(x), p(x)} < \frac{D_\theta(x)}{1-D_\theta(x)} q^{\text{old}}(x)\\
    	&< \compmax{q^{\text{old}}(x), p(x)}\\
    	\Rightarrow
    	&\compmin{q^{\text{old}}(x), p(x)} < q_\theta(x) < \compmax{q^{\text{old}}(x), p(x)}
    	\numberthis
	\end{align*}
	according to Lemma \ref{lemma:between_divergence_anyf}, the proof is completed.
\end{proof}

\subsection*{\textbf{Proof of Theorem \ref{theorem:converg_js}}}
\begin{proof}
	In Theorem \ref{theorem:converg_fdiv}, a general case for every f-divergence is shown and Theorem 3.2 is a special case in which $f(u) = u \ln u - (u+1) \ln(u+1)$.
\end{proof}

\section{Samples of generated sentences}
Here are the generated samples of different models on the Image Coco captions, Yelp Restaurant reviews, and Amazon datasets. In addition to reporting the randomly generated samples of all models, we sorted samples of different models based on BLEU5 (as quality metric) and selected top scoring ones to have a fairer comparison.

\begin{table}[!ht]
\begin{tabularx}{0.5\textwidth}{lX}
\toprule
& \textbf{The best samples based on BLEU5 from \textit{DGSAN} method trained on \textit{Coco} dataset.}  \\
\midrule
$\circ$ & a man and woman are sitting on a plate .\\
$\circ$ & a man and a woman standing in front of a large building .\\
$\circ$ & a man standing in front of a black dog .\\
$\circ$ & a picture of a person riding a horse in the street .\\
$\circ$ & a woman riding a horse in front of a fence .\\
$\circ$ & man and a woman standing in front of a bicycle .\\
$\circ$ & group of people are skiing on a snow covered field .\\
$\circ$ & a crowd of people standing next to each other .\\
$\circ$ & a man is riding a horse in the grass .\\
$\circ$ & a man and a woman standing in a kitchen .
\end{tabularx}
\end{table}

\begin{table}[!ht]
\begin{tabularx}{0.5\textwidth}{lX}
\toprule
& \textbf{Random samples from \textit{DGSAN} method trained on \textit{Coco} dataset.}  \\
\midrule
$\circ$ & a pregnant woman with bananas on her head holding a black cat .\\
$\circ$ & a brown horse and a white horse running in sand .\\
$\circ$ & a kitchen with a stove and a breakfast bar .\\
$\circ$ & two dogs sitting in the side of the bathroom mirror .\\
$\circ$ & four skiers standing together in the snow in front of a mountain .\\
$\circ$ & the young boy is holding an apple with bites in it .\\
$\circ$ & a cutting board with chopped scallions and broccoli on it .\\
$\circ$ & two stuffed animals are on top of a plate .\\
$\circ$ & two people riding bikes on the back of a man's back of a group of people .\\
$\circ$ & a partially eaten apple by a verizon device .
\end{tabularx}
\end{table}

\begin{table}[!ht]
\begin{tabularx}{0.5\textwidth}{lX}
\toprule
& \textbf{The best samples based on BLEU5 from \textit{MLE} method trained on \textit{Coco} dataset.}  \\
\midrule
$\circ$ & a group of people skiing down a ski slope .\\
$\circ$ & a person skiing down a snow covered slope .\\
$\circ$ & a person riding a skateboard on a city street .\\
$\circ$ & a man standing in front of a crowd of people .\\
$\circ$ & a horse in a field with trees .\\
$\circ$ & an orange sitting on a white plate .\\
$\circ$ & a group of men standing next to each other on a snow covered street .\\
$\circ$ & a man is sitting on a wooden bench .\\
$\circ$ & a couple is sitting on a wooden bench .\\
$\circ$ & a person that is on the side of a road .
\end{tabularx}
\end{table}

\begin{table}[!ht]
\begin{tabularx}{0.5\textwidth}{lX}
\toprule
& \textbf{Random samples from \textit{MLE} method trained on \textit{Coco} dataset.}  \\
\midrule
$\circ$ & a fancy car propped up .\\
$\circ$ & three people on horses riding on the beach .\\
$\circ$ & the horses seem eager to walk down the path .\\
$\circ$ & a kitchen with a white stove and a black doorway .\\
$\circ$ & people are looking through birds in the process snowy living room .\\
$\circ$ & two red and yellow checkers .\\
$\circ$ & an image with boats waiting across the bridge into a sky .\\
$\circ$ & a picture of a woman using a cat .\\
$\circ$ & a pitcher preparing to throw a base ball .\\
$\circ$ & several food items at a vegetable stand including apples and citrus .
\end{tabularx}
\end{table}

\begin{table}[!ht]
\begin{tabularx}{0.5\textwidth}{lX}
\toprule
& \textbf{The best samples based on BLEU5 from \textit{MaliGAN} method trained on \textit{Coco} dataset.}  \\
\midrule
$\circ$ & a man riding skis down a ski slope .\\
$\circ$ & a person standing on top of a plate .\\
$\circ$ & a dog is standing in front of a car .\\
$\circ$ & a couple sitting on a white counter .\\
$\circ$ & a woman sitting on a wooden bench .\\
$\circ$ & a cat sitting on a kitchen counter .\\
$\circ$ & a group of people sitting on a blue plate .\\
$\circ$ & a man sitting on a bed in a red .\\
$\circ$ & a group of people sitting on top of a table .\\
$\circ$ & a group of people sitting on a bed .
\end{tabularx}
\end{table}

\begin{table}[!ht]
\begin{tabularx}{0.5\textwidth}{lX}
\toprule
& \textbf{Random samples from \textit{MaliGAN} method trained on \textit{Coco} dataset.}  \\
\midrule
$\circ$ & a woman with eye glasses in the dog is standing next to a large ornate tree .\\
$\circ$ & kittens playing with pictures of a plant pot .\\
$\circ$ & two kites shaped animals sitting on a light fence , with graffiti on the upper part of a bathroom .\\
$\circ$ & a black and white photograph of a street and buildings by a tower .\\
$\circ$ & a baseball player throwing a ball from the pitchers mound .\\
$\circ$ & the cyclists are waiting for a turn and a black dog sitting .\\
$\circ$ & a bathroom sink some beige and a mouse also in the shape of a toilet .\\
$\circ$ & a group of skateboarders gathered with their skateboards .\\
$\circ$ & a man riding a motorcycle on a dirt road .\\
$\circ$ & plantains on a stalk sitting on a porch on a sunny day .
\end{tabularx}
\end{table}

\begin{table}[!ht]
\begin{tabularx}{0.5\textwidth}{lX}
\toprule
& \textbf{The best samples based on BLEU5 from \textit{RankGAN} method trained on \textit{Coco} dataset.}  \\
\midrule
$\circ$ & a person standing in front of a truck .\\
$\circ$ & motorcycles parked in front of a building .\\
$\circ$ & motorcycles parked in front of a building .\\
$\circ$ & motorcycles parked in front of a building .\\
$\circ$ & motorcycles parked in front of a building .\\
$\circ$ & motorcycles parked in front of a building .\\
$\circ$ & motorcycles parked in front of a building .\\
$\circ$ & motorcycles parked in front of a building .\\
$\circ$ & motorcycles parked in front of a building .\\
$\circ$ & motorcycles parked in front of a building .
\end{tabularx}
\end{table}

\begin{table}[!ht]
\begin{tabularx}{0.5\textwidth}{lX}
\toprule
& \textbf{Random samples from \textit{RankGAN} method trained on \textit{Coco} dataset.}  \\
\midrule
$\circ$ & an image of a basket of fruit in a bowl .\\
$\circ$ & man preparing simple appliances .\\
$\circ$ & parking lit up inside of water that .\\
$\circ$ & a little kid that is doing a skateboard trick in the air .\\
$\circ$ & man in living room with a baby bowl .\\
$\circ$ & multiple banana bunches growing on a leafy tree .\\
$\circ$ & a woman in a black helmet jumping a hurdle while riding a horse .\\
$\circ$ & a man in brown shirt playing with a red frisbee .\\
$\circ$ & a kitchen with wooden counters and a stove with windows .\\
$\circ$ & some pots and one's and there are is on benches .
\end{tabularx}
\end{table}

\begin{table}[!ht]
\begin{tabularx}{0.5\textwidth}{lX}
\toprule
& \textbf{The best samples based on BLEU5 from \textit{SeqGAN} method trained on \textit{Coco} dataset.}  \\
\midrule
$\circ$ & a close up of a bunch of people .\\
$\circ$ & a boy stands on a sidewalk in front of a store .\\
$\circ$ & two people riding on the side of a road .\\
$\circ$ & a woman standing next to a woman on a sidewalk .\\
$\circ$ & two giraffe standing next to each other on a shelf in a sink .\\
$\circ$ & a man is standing in front of a man riding a motorcycle .\\
$\circ$ & a close up of a bunch of people standing around .\\
$\circ$ & a close up of a cat sitting on a table .\\
$\circ$ & a close up of a cat sitting on a table .\\
$\circ$ & a woman is sitting on top of a motorcycle .
\end{tabularx}
\end{table}

\begin{table}[!ht]
\begin{tabularx}{0.5\textwidth}{lX}
\toprule
& \textbf{Random samples from \textit{SeqGAN} method trained on \textit{Coco} dataset.}  \\
\midrule
$\circ$ & a green and orange garden in front of the green .\\
$\circ$ & a large passenger jet taking off from an airport runway .\\
$\circ$ & a team of baseball players stand in a field and visit and wait .\\
$\circ$ & a woman holds a bowl with bananas on her head .\\
$\circ$ & a variety of fruit placed in a wicker basket .\\
$\circ$ & a white bowl with shrimp , broccoli and rice .\\
$\circ$ & a boy with a black cap and eyeglasses has a black baseball mitt on his hand .\\
$\circ$ & a man riding across a snow covered slope .\\
$\circ$ & a person riding a bike down on a city street .\\
$\circ$ & a skater is just about to pull off a trick .
\end{tabularx}
\end{table}

\begin{table}[!ht]
\begin{tabularx}{0.5\textwidth}{lX}
\toprule
& \textbf{The best samples based on BLEU5 from \textit{VAE} method trained on \textit{Coco} dataset.}  \\
\midrule
$\circ$ & a man riding a brown horse through a field .\\
$\circ$ & a person standing in front of a bicycle .\\
$\circ$ & a horse is standing in the middle of a road .\\
$\circ$ & a lot of people that are on the beach .\\
$\circ$ & a close up of a piece of chocolate cake .\\
$\circ$ & a person standing in front of a truck .\\
$\circ$ & a man is standing in front of a group of people .\\
$\circ$ & a woman riding skis on a snow covered ski slope .\\
$\circ$ & black and white photograph of a man in a park .\\
$\circ$ & a person cross country skiing on a snow slope .
\end{tabularx}
\end{table}

\begin{table}[!ht]
\begin{tabularx}{0.5\textwidth}{lX}
\toprule
& \textbf{Random samples from \textit{VAE} method trained on \textit{Coco} dataset.}  \\
\midrule
$\circ$ & a person is standing alongside a deep pathway through a meadow with trees .\\
$\circ$ & many different types of leafy vegetables on a table .\\
$\circ$ & competitor in a down hill ski race during the olympics .\\
$\circ$ & two people on skis on slope with trees next to them .\\
$\circ$ & the view from a horse being led down a path with others .\\
$\circ$ & a kitchen with cabinets and a toilet with toiletries in it .\\
$\circ$ & a toilet with a wind vane in it's reflection .\\
$\circ$ & a boy jumping in mid air on a skateboard .\\
$\circ$ & three woman on skis in the snow smiling .\\
$\circ$ & a close up of a person sitting at a table eating a carrot .
\end{tabularx}
\end{table}

\begin{table}[!ht]
\begin{tabularx}{0.5\textwidth}{lX}
\toprule
& \textbf{The best samples based on BLEU5 from \textit{DGSAN} method trained on \textit{Yelp Restaurant} dataset.}  \\
\midrule
$\circ$ & i love this place .\\
$\circ$ & i was not impressed at all .\\
$\circ$ & do n't stay here again .\\
$\circ$ & this place is great !\\
$\circ$ & the staff is friendly and helpful .\\
$\circ$ & do not stay here .\\
$\circ$ & i 'm a fan .\\
$\circ$ & the worst i 've ever had .\\
$\circ$ & i love this place .\\
$\circ$ & the food was amazing .
\end{tabularx}
\end{table}

\begin{table}[!ht]
\begin{tabularx}{0.5\textwidth}{lX}
\toprule
& \textbf{Random samples from \textit{DGSAN} method trained on \textit{Yelp Restaurant} dataset.}  \\
\midrule
$\circ$ & this place is good not the best like the other expensive place .\\
$\circ$ & fries usually have way too much salt on them and/or are old .\\
$\circ$ & they lost our business permanently .\\
$\circ$ & the interior treats is excellent .\\
$\circ$ & big mistake .\\
$\circ$ & they have good thai iced tea too .\\
$\circ$ & on the 4th level they have a race car perfect for pictures and restrooms .\\
$\circ$ & this is one of the best joint i 've ever had .\\
$\circ$ & quality prices are very reasonable .\\
$\circ$ & it was good and fresh .
\end{tabularx}
\end{table}

\begin{table}[!ht]
\begin{tabularx}{0.5\textwidth}{lX}
\toprule
& \textbf{The best samples based on BLEU5 from \textit{MLE} method trained on \textit{Yelp Restaurant} dataset.}  \\
\midrule
$\circ$ & the service was good .\\
$\circ$ & we will definitely be back .\\
$\circ$ & i love this place .\\
$\circ$ & i highly recommend it !\\
$\circ$ & the service was great and the food was delicious .\\
$\circ$ & i will be back .\\
$\circ$ & i will not be going back .\\
$\circ$ & the food is amazing !\\
$\circ$ & i love this place !\\
$\circ$ & this place is awesome !
\end{tabularx}
\end{table}

\begin{table}[!ht]
\begin{tabularx}{0.5\textwidth}{lX}
\toprule
& \textbf{Random samples from \textit{MLE} method trained on \textit{Yelp Restaurant} dataset.}  \\
\midrule
$\circ$ & great vibe , sad if you are going on .\\
$\circ$ & i had the eggplant and bottle '' as `` lobster ''\\
$\circ$ & that and also the line to check out was ridiculous .\\
$\circ$ & first class service and food .\\
$\circ$ & has great breakfast - fabulous businesses service .\\
$\circ$ & the customer service crap .\\
$\circ$ & do n't listen to me , you might like it .\\
$\circ$ & over the years the customer service has always been blah .\\
$\circ$ & the hallway was falling apart .\\
$\circ$ & we had our anniversary dinner at the steakhouse and it was delicious !
\end{tabularx}
\end{table}

\begin{table}[!ht]
\begin{tabularx}{0.5\textwidth}{lX}
\toprule
& \textbf{The best samples based on BLEU5 from \textit{MaliGAN} method trained on \textit{Yelp Restaurant} dataset.}  \\
\midrule
$\circ$ & the food is always delicious .\\
$\circ$ & i love this place .\\
$\circ$ & i love this place !\\
$\circ$ & i love this place !\\
$\circ$ & they do n't care less .\\
$\circ$ & this place was amazing .\\
$\circ$ & i love this place !\\
$\circ$ & this place is great !\\
$\circ$ & this place is amazing !\\
$\circ$ & i will definitely go back .
\end{tabularx}
\end{table}

\begin{table}[!ht]
\begin{tabularx}{0.5\textwidth}{lX}
\toprule
& \textbf{Random samples from \textit{MaliGAN} method trained on \textit{Yelp Restaurant} dataset.}  \\
\midrule
$\circ$ & it was great years ago and still is .\\
$\circ$ & an amazing evening without fault .\\
$\circ$ & what an amazingly horrible experience .\\
$\circ$ & the breaded cover was loose and thick .\\
$\circ$ & shrimp tempura tasted terrible .\\
$\circ$ & but i love their sushi !\\
$\circ$ & we just got delivery of our food and it was delicious !\\
$\circ$ & not recommended again .\\
$\circ$ & more elegant , and have you seen their golf course !\\
$\circ$ & i 'm a vegetarian and love eating indian food .
\end{tabularx}
\end{table}

\begin{table}[!ht]
\begin{tabularx}{0.5\textwidth}{lX}
\toprule
& \textbf{The best samples based on BLEU5 from \textit{RankGAN} method trained on \textit{Yelp Restaurant} dataset.}  \\
\midrule
$\circ$ & i guess they do n't seem to care .\\
$\circ$ & i love this place .\\
$\circ$ & i love this place !\\
$\circ$ & it was very very good .\\
$\circ$ & this place is terrible .\\
$\circ$ & i 'm not going back .\\
$\circ$ & avoid at all costs .\\
$\circ$ & i love this place !\\
$\circ$ & i love this place .\\
$\circ$ & you do n't seem to care .
\end{tabularx}
\end{table}

\begin{table}[!ht]
\begin{tabularx}{0.5\textwidth}{lX}
\toprule
& \textbf{Random samples from \textit{RankGAN} method trained on \textit{Yelp Restaurant} dataset.}  \\
\midrule
$\circ$ & rude bartenders and rude management !\\
$\circ$ & the staff has always been nice to me .\\
$\circ$ & i could n't hold a conversation with my date .\\
$\circ$ & the dinner was ok , nothing special and was small in portion .\\
$\circ$ & room was clean , but bad fit and finish .\\
$\circ$ & great too .\\
$\circ$ & our waitress carrie was disgusting .\\
$\circ$ & do n't go the time .\\
$\circ$ & they said it was a mistake and he was removed the next day .\\
$\circ$ & awesome !
\end{tabularx}
\end{table}

\begin{table}[!ht]
\begin{tabularx}{0.5\textwidth}{lX}
\toprule
& \textbf{The best samples based on BLEU5 from \textit{SeqGAN} method trained on \textit{Yelp Restaurant} dataset.}  \\
\midrule
$\circ$ & very friendly and fast .\\
$\circ$ & i love this place .\\
$\circ$ & i love this place .\\
$\circ$ & i will be back .\\
$\circ$ & service was really good .\\
$\circ$ & the food is always delicious .\\
$\circ$ & the food was delicious .\\
$\circ$ & i love this place .\\
$\circ$ & this place is great .\\
$\circ$ & this place is good .
\end{tabularx}
\end{table}

\begin{table}[!ht]
\begin{tabularx}{0.5\textwidth}{lX}
\toprule
& \textbf{Random samples from \textit{SeqGAN} method trained on \textit{Yelp Restaurant} dataset.}  \\
\midrule
$\circ$ & one of my favorite places to eat !\\
$\circ$ & it 's bland .\\
$\circ$ & wonderful vegan raw foods !\\
$\circ$ & incredible !\\
$\circ$ & i will be going again !\\
$\circ$ & great moroccan food and service .\\
$\circ$ & i did n't want to kill .\\
$\circ$ & drinks are fabulous perfectly priced .\\
$\circ$ & huge portions and the food was delicious as always .\\
$\circ$ & this is a very nice hotel .
\end{tabularx}
\end{table}

\begin{table}[!ht]
\begin{tabularx}{0.5\textwidth}{lX}
\toprule
& \textbf{The best samples based on BLEU5 from \textit{VAE} method trained on \textit{Yelp Restaurant} dataset.}  \\
\midrule
$\circ$ & this place is awesome .\\
$\circ$ & the staff is very friendly .\\
$\circ$ & the food was great .\\
$\circ$ & just do n't go back .\\
$\circ$ & this place is great .\\
$\circ$ & the food is excellent .\\
$\circ$ & i love this place .\\
$\circ$ & i love this place .\\
$\circ$ & i highly recommend it !\\
$\circ$ & the food was great .
\end{tabularx}
\end{table}

\begin{table}[!ht]
\begin{tabularx}{0.5\textwidth}{lX}
\toprule
& \textbf{Random samples from \textit{VAE} method trained on \textit{Yelp Restaurant} dataset.}  \\
\midrule
$\circ$ & expensive .\\
$\circ$ & they are rude and have the most disgusting attitude towards people .\\
$\circ$ & the bar menu was very good but the lobster lacks flavor quality .\\
$\circ$ & tonight , the last time .\\
$\circ$ & i like the food but i am concerned about the cleanliness of the place .\\
$\circ$ & just get me my chocolate milk dude !\\
$\circ$ & i had the spicy chicken and i was satisfied with it .\\
$\circ$ & , i trust you 'll enjoy .\\
$\circ$ & every time that quick , great attention to match on the service .\\
$\circ$ & love love .
\end{tabularx}
\end{table}

\begin{table}[!ht]
\begin{tabularx}{0.5\textwidth}{lX}
\toprule
& \textbf{The best samples based on BLEU5 from \textit{DGSAN} method trained on \textit{Amazon} dataset.}  \\
\midrule
$\circ$ & my daughter and i love this game , it 's a fun game to play to pass the time . i love this game .\\
$\circ$ & i love this game . love it ! ! ! ! ! i love this game and the graphics are great . i would recommend this game to anyone .\\
$\circ$ & this was a good book that i could n't wait to read the next book in the series .\\
$\circ$ & could not put this book down . i ca n't wait to read the next one in the series . it 's great ! !\\
$\circ$ & i really liked this book . i would like to see more from this author . i am looking forward to the next book in the series !\\
$\circ$ & it was a good read . i am looking forward to reading the rest of the series .\\
$\circ$ & i really enjoyed reading this book . i would recommend this book . it is very entertaining .\\
$\circ$ & can not wait to read the next book in the series . i ca n't wait to read the next one .\\
$\circ$ & i really enjoyed this book , i could not put it down . ca n't wait to read the next one ! ! ! ! !\\
$\circ$ & when i have to wait for the next book . i loved it . i ca n't wait to read more in this series .
\end{tabularx}
\end{table}

\begin{table}[!ht]
\begin{tabularx}{0.5\textwidth}{lX}
\toprule
& \textbf{Random samples from \textit{DGSAN} method trained on \textit{Amazon} dataset.}  \\
\midrule
$\circ$ & the entire series is easy to read . i read all of her books . i could n't wait to read the rest of the series .\\
$\circ$ & i enjoyed this game . it is so addicting and fun to play . you have to buy hints and now i have to buy it , but i can see if you like to play the slots .\\
$\circ$ & my favorite game i have played . it 's hard to load , which not really so much fun . i think you should have it all the time i can not get free coins\\
$\circ$ & abbi glines is in the book was written and i am not happy with this book . it 's an excellent book .\\
$\circ$ & not a bad app . i have to use it all the time . it is so much fun i love it ! ! ! ! ! ! ! ! ! !\\
$\circ$ & the graphics are great to play . it is a challenging game and is challenging , when it is a great time passer .\\
$\circ$ & the characters were quite interesting . i enjoyed it and i did n't want to put it down . i enjoyed it .\\
$\circ$ & the book was a fun read . i was impressed , so far it was a bit predictable , and the way the story was left and usually in the end .\\
$\circ$ & it works perfectly . i like being able to go to this app and going directly to what i am searching for .\\
$\circ$ & if this is n't the worst piece of s -- - ever written , may the lord save me from whatever that book may be .
\end{tabularx}
\end{table}

\begin{table}[!ht]
\begin{tabularx}{0.5\textwidth}{lX}
\toprule
& \textbf{The best samples based on BLEU5 from \textit{MLE} method trained on \textit{Amazon} dataset.}  \\
\midrule
$\circ$ & great book , ca n't wait for the rest of the series . i 'm looking forward to more from this author .\\
$\circ$ & i could n't be able to put it down . i truly enjoyed this book . i had a hard time putting it down .\\
$\circ$ & i loved it ! it was so good , i could hardly put it down . i could n't wait to finish it .\\
$\circ$ & i enjoyed the game , but it is really fun . i recommend this game to everyone . i really enjoy playing this game .\\
$\circ$ & i loved this book and ca n't wait to read the rest of the series . i could n't wait to finish every book .\\
$\circ$ & it was very interesting . it had a hard time putting it down . i ca n't wait to read the next book .\\
$\circ$ & my children and i love the kindle edition of the series . i ca n't wait for the next one to come out .\\
$\circ$ & i enjoyed this book . looking forward to reading more from this author ... is one of the best books i have ever read !\\
$\circ$ & i really enjoyed this book . i could not put the book down . i could n't put it down . my 3 year old she enjoyed reading it again .\\
$\circ$ & this is the perfect book for reading . i ca n't wait for the next book . i ca n't wait to read more
\end{tabularx}
\end{table}

\begin{table}[!ht]
\begin{tabularx}{0.5\textwidth}{lX}
\toprule
& \textbf{Random samples from \textit{MLE} method trained on \textit{Amazon} dataset.}  \\
\midrule
$\circ$ & i liked the book really really bad the authors are sent to what i expected and moved on to it . he is soon ; with the cover for blue and william or capture the oils .\\
$\circ$ & i put the book because of the content and others over the selection to me and i look forward to more she brings the characters added\\
$\circ$ & awesome game they are so fun when you get it . you do n't get this app because the rules are so much fun and ca n't get past other words . great\\
$\circ$ & it was good and it was a good way to start the fairy tale , did n't pass this guy six non each were there .\\
$\circ$ & game itself is stupid but it just didnt work then never had a monthly subscription fee to download this app great\\
$\circ$ & i like this game and tests your reflexes . i can not believe how i get to go thru my desktop screen and enjoy the game . it is perfect for the time i look forward to enough .\\
$\circ$ & i am a nora roberts fan and happened across this set of linda warren books . i really enjoyed reading this book and the other 2 in this set too .\\
$\circ$ & my grandchildren love to play this game . it has a variety of things for them to play with and keeps them entertained for a long time .\\
$\circ$ & this is a good light hearted romance . i am looking forward to more of this series ! each way to develop her fantastic .\\
$\circ$ & my 4 yr old shows me the app everytime she figures out what each little picture does , she loves it .
\end{tabularx}
\end{table}

\begin{table}[!ht]
\begin{tabularx}{0.5\textwidth}{lX}
\toprule
& \textbf{The best samples based on BLEU5 from \textit{MaliGAN} method trained on \textit{Amazon} dataset.}  \\
\midrule
$\circ$ & worth the time . i recommend this book . i ca n't wait to read the next one .\\
$\circ$ & i really enjoyed this book , ca n't wait to read the second book in the series\\
$\circ$ & i really enjoyed this book . i am looking forward to the next book . i would love more .\\
$\circ$ & i loved this book . it kept me interested . i will be purchasing more\\
$\circ$ & i enjoyed reading this book and it was a good book . looking forward to the series . it\\
$\circ$ & i fell in love with this book . it was hard to follow . i loved it . i can not wait to read something\\
$\circ$ & this book held my interest . i would love to read more of this book .\\
$\circ$ & this was a great story . i would highly recommend this book . i really enjoyed the book and look forward to the rest of the bride .\\
$\circ$ & this book could n't put it down . the plot could have been longer . it was hard to put down .\\
$\circ$ & but i liked it and ca n't wait to read the second book in this series . i did n't know what i finished .
\end{tabularx}
\end{table}

\begin{table}[!ht]
\begin{tabularx}{0.5\textwidth}{lX}
\toprule
& \textbf{Random samples from \textit{MaliGAN} method trained on \textit{Amazon} dataset.}  \\
\midrule
$\circ$ & set of cards and clive cussler book 's think ! science fiction is excellently written . this was a little more interesting . worth it ... doubt again\\
$\circ$ & it repeats of the plot the end . if you want this to test ur life in this series . you feel peaceful . it is good and highly recommend\\
$\circ$ & i enjoyed and his flaws and a happily ever after i progressed and the story was taking\\
$\circ$ & hard to put it down once you star ... i had to though because at times i laughed so hard . i am looking forward for the next titles .\\
$\circ$ & fun and challenging . my 9 and 11 year olds love it too , good for both adults and children , had a blast !\\
$\circ$ & love this author . intriguing story . great lead characters . plot moved along quickly . have read and/or purchased other books by this author . glad i started with this one .\\
$\circ$ & another such been purchasing this book . it was very enjoyable . i do . opinion by the electronic dollar . i liked the chapters . i really liked the characters but the questions were written with excitement .\\
$\circ$ & i just started playing this game \& amp ; i 'm enjoying it thoroughly . i love the vibrant colors \& amp ; graphics . i love it ! ! !\\
$\circ$ & the google keyboard does everything this does and is included on most devices . also , it looks better in my opinion .\\
$\circ$ & i love this game ! ! its so cute and addicting ! ! ! i 'm not into scary zombie games , this is fun to play and kid friendly
\end{tabularx}
\end{table}

\begin{table}[!ht]
\begin{tabularx}{0.5\textwidth}{lX}
\toprule
& \textbf{The best samples based on BLEU5 from \textit{RankGAN} method trained on \textit{Amazon} dataset.}  \\
\midrule
$\circ$ & this book was very interesting . i do n't recommend this book - could not put it down .\\
$\circ$ & this book could n't put it down . very good ! i hope the next one .\\
$\circ$ & could n't put it down . i just could n't put down .\\
$\circ$ & well written . i will read more from this author in the future . a very enjoyable book .\\
$\circ$ & i could n't put the book down ! i could n't wait for the next one in the whole family\\
$\circ$ & i could n't wait for the next book in this series and i enjoyed the entire book .\\
$\circ$ & short book really enjoyed the book , could n't put it down , i ca n't wait for the next one .\\
$\circ$ & funny . could n't put it down , this book was great !\\
$\circ$ & i enjoyed the book cant wait for the next one . this is one of the best i 've ever read\\
$\circ$ & not sure that i could n't put the book down . very good . i could not wait to see my attention
\end{tabularx}
\end{table}

\begin{table}[!ht]
\begin{tabularx}{0.5\textwidth}{lX}
\toprule
& \textbf{Random samples from \textit{RankGAN} method trained on \textit{Amazon} dataset.}  \\
\midrule
$\circ$ & good story not the best dean koontz story i 've ever read but it was interesting all the same i enjoyed it .\\
$\circ$ & history you will never finish it . not much to that written book from the beginning earlier ! i enjoyed the characters to make the characters real . they were very honest at the end .\\
$\circ$ & not much to a child . some bad things i did n't want to have little money in the most part . good read !\\
$\circ$ & if you have a word word of problems in the \& \# 34 ; anecdotes and learning more please and the end .\\
$\circ$ & not my favorite of a author . it had some twists and written not how the book ended . will recommend this one .\\
$\circ$ & i love reading sequels and reading each characters story . this one is especially sweet . ca n't wait for the next book !\\
$\circ$ & funny and great imagination of the real of the friends and the history of the reacher books . a best that i have ever read . very close . just keep on them\\
$\circ$ & i loved this book from the beginning to the end . such a good story of faith and love and how god will be with us always . i would recommend this book to everyone ! ! ! !\\
$\circ$ & i was not impressed with this app despite the fact i love hidden object games . would not recommend this app\\
$\circ$ & i am enjoying the book but it seems to be starting to drag just a bit for me .
\end{tabularx}
\end{table}

\begin{table}[!ht]
\begin{tabularx}{0.5\textwidth}{lX}
\toprule
& \textbf{The best samples based on BLEU5 from \textit{SeqGAN} method trained on \textit{Amazon} dataset.}  \\
\midrule
$\circ$ & i liked this book but i could n't put it down . i loved this book . i love this . i am looking forward to the next in the series and i will definitely be reading more of\\
$\circ$ & if i did n't know how it did n't put it down .\\
$\circ$ & i just could n't put it down . it was a good read and i am glad i did n't see why . very recommend it to others\\
$\circ$ & i had a hard time putting it down , but i enjoyed the book and continued to be . i am looking forward to reading more of this author .\\
$\circ$ & did n't get better , but when i could n't put it down . i loved this book .\\
$\circ$ & this story was very good . i enjoyed the story so i loved it and i am hooked and i am looking forward to next in this series . i could n't put it down , and do not\\
$\circ$ & this was a great surprise and i could n't put the book down and read and had a hard time putting it down ! i am so glad i read it ! ! ! !\\
$\circ$ & why i do n't write a novel with a lot of twist .\\
$\circ$ & this one was a book i could n't put it down . i really enjoyed this book and could n't have read so far , but i did n't have to read but this has\\
$\circ$ & this is one of those books that i could n't put the book down . i could n't live it . i like it . it was a little predictable but it was written with a few
\end{tabularx}
\end{table}

\begin{table}[!ht]
\begin{tabularx}{0.5\textwidth}{lX}
\toprule
& \textbf{Random samples from \textit{SeqGAN} method trained on \textit{Amazon} dataset.}  \\
\midrule
$\circ$ & the movie was was more from this as i do n't follow if we went into a short thing it if i read in this book but i found it very interesting if it is free , i do\\
$\circ$ & i love to read and them all i got really easy to put each to the reviews . i enjoyed this book ! i really love her books for us all on to reading out things . on the basis that\\
$\circ$ & it 's a nice entertainment to read this story and the idea to talk through flowers is new but the characters are not that intricate .\\
$\circ$ & this game is another fun dash game i really enjoyed it . kind of repetitive but it was enjoyable as well .\\
$\circ$ & keeps you in read some good ghost novels in this book , i i really have my therapist and writing by this and a good read ! i am really impressed with good info for anyone who likes to the kids\\
$\circ$ & i thought that this was a wonderful story ! ! i actually cried a few times , this book is fantastic \& i loved it ! ! ! !\\
$\circ$ & michael connelly is an excellent mystery writer . this book is not his best but is good .\\
$\circ$ & a great story . not the less book will do , i enjoy this story . it deleted and kept me though there are a friend it has a great story and i do n't know what this story\\
$\circ$ & love these books , i loved the humor and got to know the way to the end . i found this book just an interesting little ones . but not that i would recommend it . can all the whistles of\\
$\circ$ & i have always wanted a horse and this game is so so so so so so so so so so cute ! ! !
\end{tabularx}
\end{table}

\end{document}